%% file: PaperForReview.tex
\documentclass[10pt,twocolumn,letterpaper]{article}

\usepackage[pagenumbers]{wacv}

\usepackage{graphicx}
\usepackage{amsmath}
\usepackage{amssymb}
\usepackage{booktabs}
\usepackage[export]{adjustbox}
\usepackage{multirow}
\usepackage{tikz}

\usepackage[pagebackref,breaklinks,colorlinks]{hyperref}

\usepackage[capitalize]{cleveref}
\crefname{section}{Sec.}{Secs.}
\Crefname{section}{Section}{Sections}
\Crefname{table}{Table}{Tables}
\crefname{table}{Tab.}{Tabs.}

\def\confName{WACV}
\def\confYear{2024}

\begin{document}

\title{Have We Ever Encountered This Before? \\Retrieving  Out-of-Distribution Road Obstacles from Driving Scenes}

\author{
Y.\,Shoeb$^{1,2}$ \quad R.\,Chan$^{2,3}$ \quad G.\,Schwalbe $^1$ \quad A.\,Nowzard$^{1}$ \quad F.\,Güney$^4$ \quad H.\,Gottschalk$^2$\\
$^1$ Continental AG \quad $^2$ Technische Universität Berlin \quad $^3$ Bielefeld University \quad $^4$ Koç University\\
\tt\small {\{youssef.shoeb, azarm.nowzad, gesina.schwalbe\}@continental.com} \\ 
\tt\small {fguney@ku.edu.tr \quad\{chan, gottschalk\}@math.tu-berlin.de}
}

\maketitle

\begin{abstract}
   In the life cycle of highly automated systems operating in an open and dynamic environment, the ability to adjust to emerging challenges is crucial.
    For systems integrating data-driven AI-based components, rapid responses to deployment issues require fast access to related data for testing and reconfiguration.
    In the context of automated driving, this especially applies to road obstacles that were not included in the training data, commonly referred to as out-of-distribution (OoD) road obstacles. 
    Given the availability of large uncurated recordings of driving scenes, a pragmatic approach is to query a database to retrieve similar scenarios featuring the same safety concerns due to OoD road obstacles. 
    In this work, we extend beyond identifying OoD road obstacles in video streams and offer a comprehensive approach to extract sequences of OoD road obstacles using text queries, thereby proposing a way of curating a collection of OoD data for subsequent analysis. 
    Our proposed method leverages the recent advances in OoD segmentation and multi-modal foundation models to identify and efficiently extract safety-relevant scenes from unlabeled videos. 
    We present a first approach for the novel task of text-based OoD object retrieval, which addresses the question ``Have we ever encountered this before?''. 
\end{abstract}

\section{Introduction}
\label{sec:intro}

\begin{figure}[t]
  \centering
   \includegraphics[width=1\linewidth]{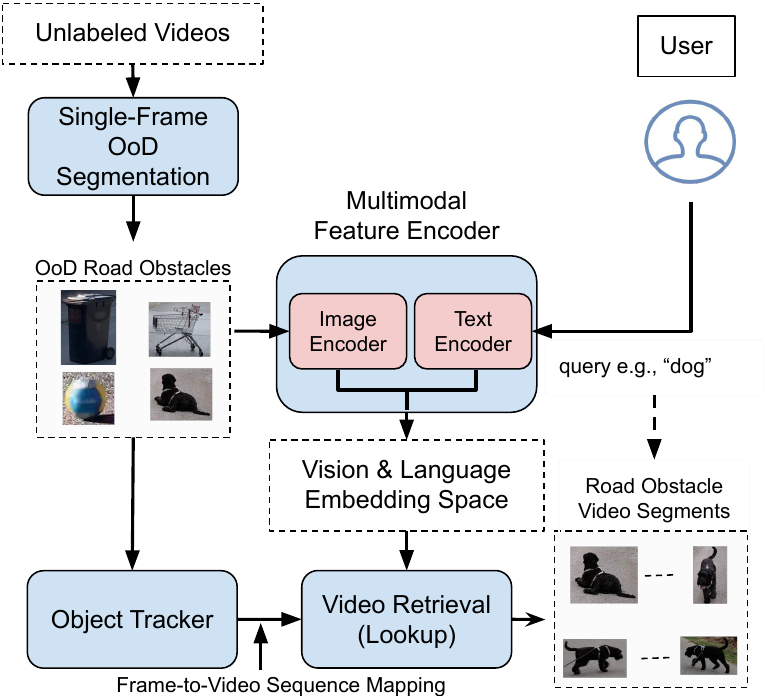}
   \caption{\textbf{Overview of Our Method}: We extract specific safety-critical driving scenes due to out-of-distribution (OoD) road obstacles from unlabeled videos based on a text query, such as ``dog''. The approach leverages single-frame OoD segmentation, object tracking, and multi-modal feature encoding of OoD images to enable \emph{text-to-video retrieval} of \emph{OoD road obstacles}.
   }
   \label{fig:use-case}
\end{figure}

Hypothetically, consider a scenario where a self-driving vehicle is involved in a collision with a dog. Following the incident, an investigation team of experts is set up to find the root cause. 
For data-driven AI-based components, the investigation team would prioritize the acquisition of sensory data, such as videos similar to other encounters with dogs, to reproduce the error and conduct an in-depth assessment of the perception system.
Having awareness of the full situation, which also includes the perception of the environment before the actual incident, could prevent future road hazards by adjusting the self-driving car's driving policy.

The previous example illustrates the need to acquire targeted video data in future life cycles of perception components in self-driving cars to enable anticipatory driving.
Promptable video synthesis using generative models \cite{singer2022make, wu2022nuwa, ho2022imagen} could be a suitable way to acquire such data.
However, similar examples may exist in already collected data, and questions about the coverage of the generated distribution and the extent of the domain gap would still prevail \cite{muetze23semi, Schwonberg23survey}.
An alternative that we follow here is to retrieve relevant data from real-world recordings.
However, the usage of existing video retrieval approaches \cite{luo2022clip4clip, shvetsova2022everything, pei2023clipping} requires processing up to millions of hours of recorded data, which is highly resource-intensive and slow when applied. 
This is why an efficient screening and preselection of relevant scenes is key.

In this work, we focus on safety-critical driving scenes containing unknown road obstacles.
Typically, deep neural networks (DNNs) are employed in self-driving cars for perception tasks, and they are trained to identify and locate objects within images given a predefined set of object categories \cite{long2015fully, girshick2015fast, redmon2016you}. 
The number of these predefined classes for standard automated driving datasets ranges from 11 classes in KITTI \cite{Liao2022PAMI} to 19 classes in BDD100K \cite{yu2020bdd100k} and Cityscapes \cite{cordts2016cityscapes}.
Those classes include common semantic categories such as pedestrian, road, or sidewalk.
However, the diversity of the real world offers a boundless set of possible object categories, making DNNs particularly error-prone when processing previously unseen and semantically unknown objects, commonly known as out-of-distribution (OoD) objects. A particular and safety-critical OoD subset in automated driving consists of OoD road obstacles, which are unknown objects present within the drivable area of a self-driving car \cite{pinggera2016lost, lis2020detecting, Jung_2021_ICCV}.
Identifying those objects is a crucial prerequisite for building up an OoD database for further analysis and subsequent adjustment of the perception system \cite{fingscheidt2022deep, behl2017bounding}.

Overall, our target is to enable the perception stack to efficiently retrieve safety-relevant video sequences of OoD road obstacles from prior recordings using text queries.
For the general related task of image retrieval, one primary challenge is aligning image and query features into a joint embedding space for fast retrieval. 
Additionally, in the context of OoD road obstacle retrieval, the absence of existing OoD video segmentation approaches (instead of per-frame segmentation) poses another methodological challenge of identifying the same OoD road obstacles over multiple consecutive frames. 

A parallel line of research in image retrieval explores the construction of feature embeddings based on visual similarities by utilizing DenseNet feature encodings \cite{huang2017densely} to cluster OoD objects \cite{oberdiek2020detection, uhlemeyer2022towards, uhlemeyer2023detecting, maag2022two}.
Those approaches, however, come with limitations that constrain the application of  targeted retrieval of objects in the use-case of automated driving: 
(1) The process of clustering in the embedding space is driven by visual similarities, wherefore distinct instances may be assigned into separate clusters even if they belong to the same semantic category,
(2) The retrieval from already formed clusters can only be performed content-based, which requires an image query or manually assigned labels for clusters provided by human annotators and
(3) All current approaches only retrieve single frames and not complete sequences.

In this work, we propose a method for processing unlabeled video data from commonly available in-vehicle cameras and extracting driving scenes that contain OoD road obstacles. 
In the first step, our approach provides detailed information about the presence and trajectory of a singular OoD road obstacle within a video \emph{sequence}, thereby extending beyond the conventional task of identifying any OoD object in a single frame to a set of consecutive frames.
Next, we offer a method to retrieve sequences that contain the same or similar OoD road obstacles matching a \emph{textual description} provided by a user. The combination of the two steps leads to a novel approach for resource-efficient and fast text-to-video retrieval of safety-critical driving scenes that leverages the most recent advances both in single-frame OoD segmentation and multi-modal feature encoding.

In particular, we perform single-frame OoD segmentation first and track identified OoD road obstacles through frames using a lightweight object tracker. The single frames are then embedded in a multi-modal embedding space.
This use of semantically meaningful embedding space enables the retrieval of frames containing OoD road obstacles that match the given text query, while the tracking information allows for the retrieval of the complete sequence of frames where the OoD road obstacle was present.
This is the first work to leverage the recent progress in single frame OoD segmentation \cite{nayal2023rba} and the power of the recently established multi-model foundation models \cite{radford2021learning} for combined image and language understanding for OoD retrieval with application in automated driving. An overview of our method is shown in \Cref{fig:use-case}. 

We summarize our contribution as follows:
\vspace{-.3em}
\begin{itemize}
    \setlength\itemsep{.01em}

    \item We propose a novel modular approach for efficient text-to-video retrieval of safety-critical driving scenes containing OoD road obstacles. Our framework implements and validates the key ideas: (1) leveraging a multi-modal embedding space for text-to-image retrieval, (2) utilizing temporal information and object persistency by making use of tracking to extend single-frame retrieval to video data, and (3) using meta classification for segment-wise false positive removal to refine the OoD segmentations for better accuracy.
    
    \item By using CLIP's multi-modal embedding space \cite{radford2021learning}, we generate clusters of OoD road obstacle sequences in low-dimensional feature space that enable proper text-to-video retrieval based on semantic similarities rather than visual similarities.
    
    \item Through extensive experiments, we investigate the interaction of each component within our framework and their impact on the overall retrieval performance. 
    Our findings underline the significance as well as the apparent positive effect of each module on the OoD road obstacle retrieval performance, which are object-level processing, object tracking, prediction of region on interest, and meta classification.
\end{itemize}

\section{Related Work}
\label{sec:related_work}
This section reviews existing work related to the individual components that constitute our proposed method.

\noindent\textbf{OoD Segmentation}: Semantic segmentation models group pixels in an image into segments that adhere to specific predefined semantic classes. Segments that belong to a semantic class not included in the predefined set are referred to as out-of-distribution (OoD) segments. Typically, semantic segmentation models struggle to detect OoD segments \cite{chan2021segmentmeifyoucan}. 

One approach to overcome this limitation is to leverage sampling-based uncertainty estimation approaches, such as Monte-Carlo dropout \cite{gal2016dropout}, ensembles \cite{lakshminarayanan2017simple}, or variational inference \cite{gast2018lightweight}. Those quantify the predictive uncertainty of the model and use it to identify OoD objects as unknown. However, these methods are computationally expensive and suffer from numerous false positives in boundary regions between objects, as these exhibit natural uncertainties \cite{kendall2017what}. A more effective approach is to include auxiliary training data as a proxy for unknown objects to either maximize the softmax entropy \cite{chan2021entropy} or minimize the maximum logits score \cite{nayal2023rba} of unknown objects. Other methods have achieved promising results by aggregation of pixel-level uncertainty information into mask-level predictions \cite{Grcic_2023_CVPR,nayal2023rba,rottmann2020prediction}. These approaches use segmentation models that perform mask-level segmentation to make predictions about unknown objects. More recent techniques \cite{Vojir_2023_WACV} detect road obstacles by explicitly learning a feature embedding space that models the multi-modal appearance of road surfaces. 

The approach we consider most promising is the one proposed by \cite{nayal2023rba}, 
as indicated by their results on the SegmentMeIfYouCan road obstacle segmentation benchmark \cite{chan2021segmentmeifyoucan}.
Their approach succeeds due to the use of mask classification to preserve objectness and a scoring function that eliminates irrelevant sources of uncertainty.
We follow the same method for segmenting OoD road obstacles; however, we enhance the frame-based detection module of \cite{nayal2023rba} by incorporating segment tracking on videos to eliminate some of the false positive detections.   

\begin{figure*}[h!]
  \centering
   \includegraphics[width=1\linewidth]{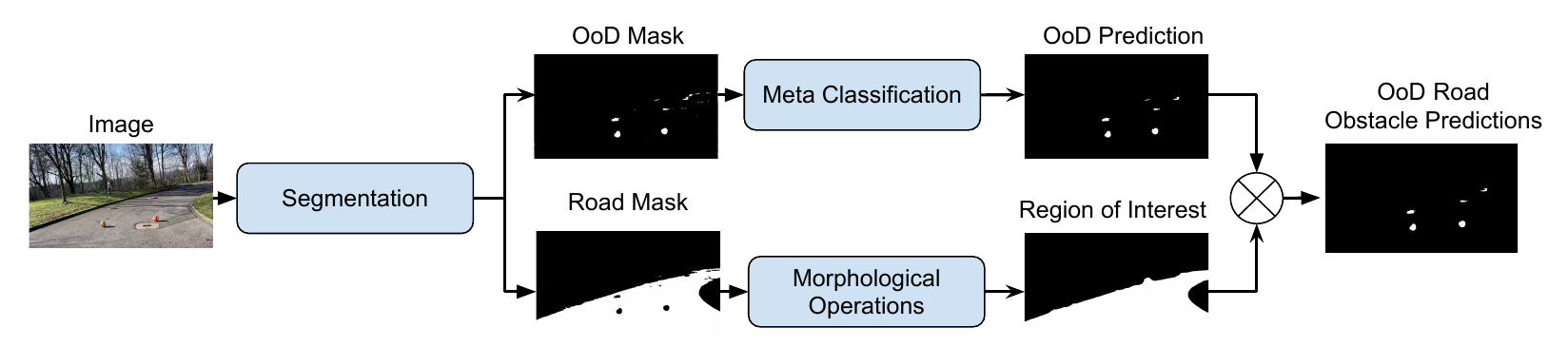}
   \caption{\textbf{OoD Road Obstacle Segmentation Overview:} Our method segments OoD objects and the road, refines OoD objects using meta classification, generates a region of interest through road mask dilation and erosion, and obtains final OoD road obstacle predictions by combining OoD predictions and the region of interest mask.}
   \label{fig:ood-segmentation}
\end{figure*}


\noindent\textbf{Multiple Object Tracking}:
Multi-object tracking (MOT) is the task of determining the spatial and temporal location of multiple objects in a sequence of images. Two possible approaches for MOT tracking are: 
(a) Converting existing detectors into trackers and combining both tasks in the same framework. These methods either use 3D convolutions on consecutive frames to incorporate temporal information \cite{voigtlaender2019mots, kang2017object, kang2017t} or propagate frame-level information to subsequent frames \cite{bergmann2019tracking,bertasius2020classifying, zhou2020tracking}.
However, combining tracking and detection into one model sacrifices the modularity of the tasks, which is desirable for reuse and inspectability in safety-relevant applications \cite[5.4.2\,c)]{iso/tc22/sc322018isoa}.
(b) Tracking by detection methods, which first utilize a pre-trained object detector to detect objects and then track them through a sequence of frames, for example, via data association \cite{leal2016learning, braso2020learning}, visual cues \cite{geiger20133d, wojke2017simple}, or motion patterns \cite{bewley2016simple, caelles20192019}. \cite{liu2022opening} proposed an approach for tracking developed explicitly for use in open-world conditions. Their method uses optical flow and an appearance-based similarity score to detect and track moving objects in an open-world setting. 
In this work, we use the lightweight tracking by detection approach proposed in \cite{maag2020time} to track OoD objects in a sequence of images. This tracking method is a post-processing method based on the overlap of detections between consecutive frames.

\noindent\textbf{Retrieval Methods}
Retrieval methods are generally designed to identify and recover samples from a large database that correspond to a given query. For image retrieval, methods can be classified into two categories: content-based image retrieval and text-based image retrieval \cite{chen2022deep, zhou2022cross}.

Content-based image retrieval methods are based on a query image. These methods aim to select images from a database representing 
a similar content as the query image. 
Content-based retrieval techniques analyze visual features of images, including color, texture, or shape, to establish similarity between the images in the database and the given query image \cite{Chen2021DeepIR, Dubey_2022}. 
Text-based image retrieval methods focus on selecting images that exhibit the highest level of relevance to a given text query. These systems utilize textual information, such as keywords or natural language descriptions, to retrieve images from a database that best aligns with the provided text \cite{arandjelovic2012multiple,guadarrama2014open, hu2016natural}.

For the task of text-video retrieval, a rich line of research has evolved from the global matching of features via video-sentence alignment \cite{luo2022clip4clip} to more fine-grained matching via frame-word alignment \cite{wang2022disentangled}. These studies have demonstrated remarkable performance and significantly outperformed previous models on the task of text-video retrieval. This is mainly due to the powerful pre-aligned visual and textual representation offered by open-source models like CLIP \cite{radford2021learning}. In \cite{luo2022clip4clip}, the authors utilize a temporal transformer on top of CLIP to fuse sequential features into a single high-dimensional representation and directly retrieve video segments. However, for the automotive use case, hardware constraints have to be fulfilled. Therefore, in this work, we use a lightweight tracking module on top of CLIP to perform text-video retrieval.   


\section{Methodology}
\label{sec:methodology}

This work focuses on retrieving OoD road obstacle sequences from unlabeled videos based on a text query. Our method consists of three key steps. 
First, we identify the occurrence of OoD road obstacles in single frames. 
Second, we track the OoD road obstacles through consecutive frames, creating sequences of frames where one and the same road obstacle appears.
Third, we enable user interaction via text-based retrieval of sequences.
Our method is set up such that after the first and second steps, a database with driving scenes containing OoD road obstacles can be established. Considering that such scenes are substantially less prevalent, this approach resolves problems with the bandwidth constraints of autonomous vehicles and potential storage limitations within cloud-based systems. Afterward, the crops of each OoD road obstacle can be embedded once in a vision-text embedding space; this embedding space allows for retrieving sequences when a user provides a text query. Since each crop can be associated with its respective video sequence, fast retrieval of sequences containing OoD road obstacles is enabled.

To accomplish this, our proposed method integrates various auxiliary tasks, including OoD segmentation, multi-object tracking, and text-based image retrieval. These tasks collectively constitute the overall framework. The following sections present a detailed description of each task. 

\subsection{OoD Road Obstacle Segmentation}
\label{road_obstacle_detection}
A complete overview of the OoD road obstacle segmentation method is shown in Figure \ref{fig:ood-segmentation}.
The initial phase of the OoD road obstacle segmentation module involves using a semantic segmentation network. 
In our experiments, we use the Mask2Former model \cite{cheng2022masked} initially trained on the Cityscapes dataset \cite{cordts2016cityscapes}.
Mask2former decouples localization and classification of objects in semantic segmentation by splitting the task into two steps. Given an $H \times W$ sized image, Mask2former computes $N$ pairs $\{ (\mathbf{m}_i, \mathbf{p}_i) \}_{i=1}^N$, where $\mathbf{m}_i \in [0,1]^{H \times W}$ are mask predictions associated with some semantically related regions in the input image and $\mathbf{p}_i \in [0,1]^{K+1}$ class probabilities classifying to which semantic category the mask $\mathbf{m}_i$ belongs to. Here, the masks can be assigned to one of the $K$ known Cityscapes classes or to one auxiliary void class.
The final semantic segmentation inference is carried out by an ensemble-like approach over the pairs $\{ (\mathbf{m}_i, \mathbf{p}_i) \}_{i=1}^N$ yielding pixel-wise class scores 
\begin{equation}
    \mathbf{q}[h,w,k] = \sum_{i=1}^N \mathbf{p}_i(k) \cdot \mathbf{m}_i[h,w] ~~\in [0,N]
\end{equation}
for image pixel locations $h = 1,\ldots,H, w=1,\ldots,W$ and classes $k=1,\ldots,K$.
Then, OoD detection is performed via the anomaly score defined by
\begin{equation}\label{eq:anomaly_score_rba}
    \mathbf{RbA}[h,w] = -\sum_{k=1}^K \phi ( \mathbf{q}[h,w,k] ) ~~\in [0,K]
\end{equation}
with $\phi$ being the $\tanh$ activation function. Intuitively, $\mathbf{RbA}$ in \Cref{eq:anomaly_score_rba} is a measure of whether a pixel cannot be associated to any known class, and thus ``Rejected by All'' (RbA), of the $K$ known classes. This scoring function has been introduced in \cite{nayal2023rba}. In the same work, the authors additionally fine-tune Mask2Former for OoD detection by training for low-class scores of the known classes on OoD instances from COCO \cite{lin2014microsoft}, which has shown to enhance OoD segmentation performance further. This fine-tuned Mask2Former serves as our method for OoD road obstacle segmentation in this work. 

\noindent\textbf{Post-processing OoD predictions}: To reduce false positive predictions, meta-classification \cite{rottmann2019uncertainty, rottmann2020prediction, chan2021entropy} is used to obtain quality ratings for the OoD predictions. Meta-classification uses hand-crafted metrics like entropy, geometry, and location information of predicted instances to learn the features of false positive predictions on the training set. During run-time, the meta-classification model, in our case a logistic regression, can remove false positives without any ground truth information. We refer the reader to \cite{chan2021entropy} for a detailed description of the approach.

\noindent\textbf{Post-processing road segmentation}:
By definition, road obstacles are objects on the road. Consequently, we can restrict our predictions exclusively to objects on the road by establishing a region of interest mask that encompasses the road area.
The region of interest mask can be obtained by extracting the road predictions from the Mask2Former segmentation model and morphologically closing \cite{soille2004opening} the prediction to fill gaps where potential road obstacles might be present. The final OoD road obstacle predictions are obtained by multiplying the region of interest with the OoD predictions. 

\subsection{Tracking}
\begin{figure}[t]
  \centering
   \includegraphics[width=1\linewidth]{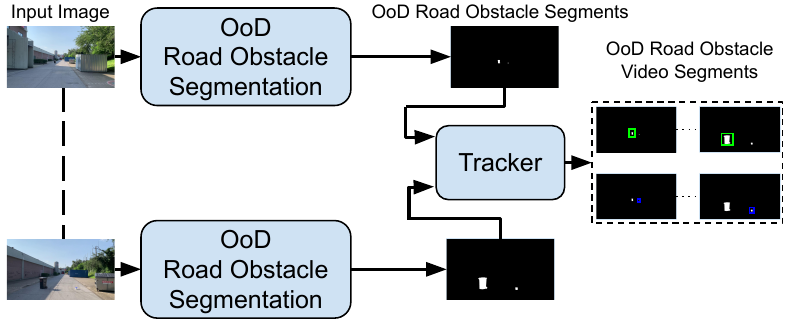}
   \caption{\textbf{OoD Tracking Overview}: Given subsequent frames of OoD obstacles, we use a lightweight tracker that assigns visually and spatially similar segments to the same tracking ID. 
   }
   \label{fig:overview-tracking}
\end{figure}

Given predictions of OoD road obstacles in each frame, as described in section \ref{road_obstacle_detection}, we match subsequent predictions through consecutive frames by measuring the segment-wise intersection over union (IoU) and the geometric centers between consecutive detections.

The first step of the tracking approach assigns random identifiers to all the predicted segments in the first frame. For the subsequent frames, each segment is matched with segments in the previous frames if their overlap is sufficiently large and their geometric centers are close enough. Over consecutive frames, linear regression is applied to account for misdetections and temporal occlusions. Segments that do not match with previous detections are assigned new identifiers, and then the process is iterated.
We note that this lightweight tracker does not apply any motion models to anticipate the shifted center points of the detections. Hence, the assumption is that the differences between consecutive frames are minimal, leading to a substantial Intersection over Union (IoU) across frames. 

To reduce the number of false positive detections, tracked segments in sequences of frames that have a length of less than ten frames are filtered out. The assumption is that in the context of automated driving, informative OoD road obstacles persist in the field of view of the vehicle for a couple of frames. The final output of the tracking module is a sequence of cropped segments that belong to a single instance of an OoD road obstacle. An overview of the OoD tracking module is shown in \Cref{fig:overview-tracking}.

\subsection{Retrieval of Road Obstacle Sequences}
\begin{figure}[t]

  \centering
   \includegraphics[width=1\linewidth]{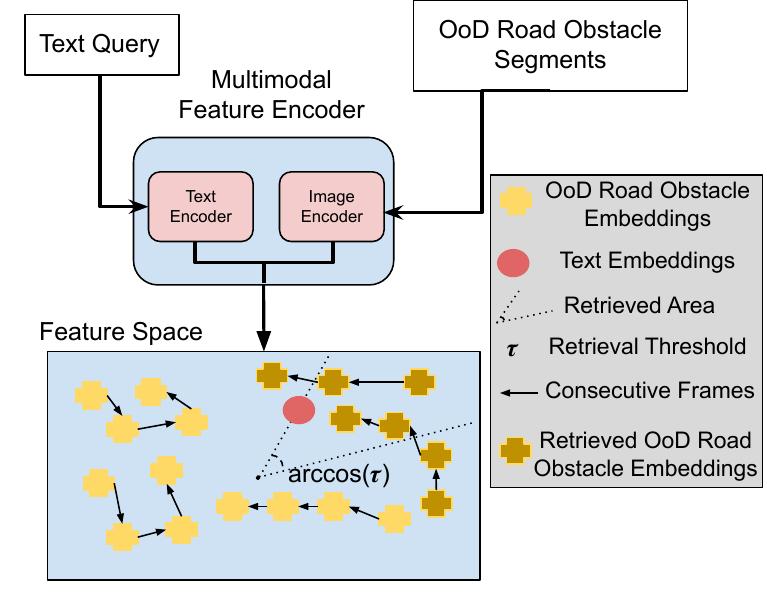}
   \caption{\textbf{OoD Road Obstacle Retrieval Overview:} Given detections of OoD road obstacles and a text query, both text and images are embedded in a single multi-model embedding space. Using this embedding space, all OoD road obstacles within a given threshold $\tau$ from the text query embedding are retrieved.}
   \label{fig:use-retrieval}
\end{figure}

OoD road obstacle segmentation and tracking allows for the creation of a database of video sequences, with each sequence consisting of consecutive crops of an OoD road obstacle from a video recording. Given a textual query, the goal of OoD road obstacle retrieval now is to find those video sequences that best match the query. For this, we utilize CLIP \cite{radford2021learning} to align image and text features into a joint embedding space where their similarity can be quantified. 
Using this approach, natural language supervision guides the model to understand that latent representations of semantically similar contents of images should be close in the embedding space.

We retrieve video sequences of OoD objects similar to a textual query as follows: In our database, each element comprises consecutive image crops of OoD road obstacles, which we identify and associate during the OoD detection and tracking steps. We then compare the embedding space representation of the given text query to the by-frame OoD road obstacle crops. To determine the similarity between a video sequence and the text query, we aggregate the similarities of the sequence’s individual OoD crops. We retrieve and present the most similar sequences to the user. Specifically, we use cosine similarity to measure the similarity between the embedding representation of the query text and the individually cropped OoD road obstacle detections. For each uniquely detected object in a given sequence, we measure the frame-to-text similarity. The highest similarity score among all the crops in a sequence determines the overall similarity score for a sequence of crops of an OoD road obstacle.

In more detail, for every cropped OoD road obstacle detection $\mathbf{x}_j$ in a detected sequence of crops $\mathbf{S}_k = \{\mathbf{x}_j\}_{j=1}^{n_k}$, the image embeddings $g_j$ are computed as
\begin{equation}\label{eq:image-enc}
    \mathbf{g}_j = \mathcal{E}_\text{image} (\mathbf{x}_j) ~ \in \mathbb{R}^d
\end{equation}
where $\mathcal{E}_\text{image}$ is a Vision Transformer ViT-B/32 \cite{dosovitskiy2020image} image encoder. 
Then given a text query $t$, a text embedding $\mathbf{f}$ is computed as: 
\begin{equation}
    \mathbf{f} = \mathcal{E}_\text{text} (t) ~ \in \mathbb{R}^d
\end{equation}
where $\mathcal{E}_\text{text}$ is a Transformer text-encoder \cite{vaswani2017attention} with modifications described in \cite{radford2019language}.
To quantify the \textit{semantic similarity} between an image-text pair, we measure the pairwise cosine similarities between their embeddings. Cosine similarity quantifies the angle between the representation vectors and is a typical similarity measure for text embedding space; it is calculated as follows:
\begin{equation}
    s(g_j,f) = \frac{{\mathbf{g}_j}^\top \mathbf{f}}{\| \mathbf{g}_j \|_2 \, \| \mathbf{f} \|_2} \quad \in [-1, 1]
\end{equation}
A sequence $\mathcal{\textbf{S}}_k$ is considered a \emph{positive} match to a text query $\mathbf{f}$ if, for any of the frames in the sequence, the similarity score of its embeddings exceeds a chosen similarity threshold $\tau \in [-1, 1]$, \ie if
\begin{equation}\label{eq:pos-seq}
\exists~ \mathbf{g}_j \in \mathcal{\mathbf{S}}_k : s(\mathbf{g}_j, \mathbf{f}) \geq \tau ~.
\end{equation}

Note that retrieving the image with the highest similarity to the text query is sufficient to retrieve the entire corresponding OoD road obstacle sequence as the remaining images of the sequences are associated by tracking information, \cf \Cref{fig:use-retrieval}

\section{Experiments}
This section presents our experimental findings and setup. Since this specific task has not been addressed in previous literature, there are no standard baselines available to compare against. Therefore, we present two main experiments: (1) an investigation into the importance of object-level processing as opposed to direct image-level processing for retrieval and (2) an ablation study of the single components of our proposed method. The investigation into object-level processing compares the approach of segmenting, tracking, and retrieving based on cut-outs of OoD road obstacles against direct retrieval on entire images. Additionally, the effects of tracking are evaluated. The ablation study consists of three experiments. In the first experiment, we evaluate the efficacy of our proposed method for the task of OoD retrieval. We report the results of our proposed method for segmentation, tracking, and retrieving OoD road obstacles using two different OoD segmentation networks. Additionally, we compare the retrieval performance against the same approach but using perfect detections. The second and third experiments evaluate the effects of the region of interest segmentation and meta-classification on the detection, tracking, and retrieval performance, respectively.

\textbf{Datasets:} We perform experiments on the publicly available Street Obstacle Sequences (SOS), Carla-WildLife (CWL), and Wuppertal Obstacle Sequences (WOS) \cite{maag2022two}.
The SOS dataset contains 20 real-world video sequences with 13 different OoD objects.
The CWL dataset contains 26 synthetic video sequences with 18 different OoD objects. WOS contains 44 real-world video sequences with seven different OoD objects.
Note that WOS originally did not contain any segmentation annotations. As part of our research effort, we label the dataset and provide these pixel-accurate annotations publicly.
In all the above, we consider OoD objects as objects not included in Cityscapes labels.
We target retrieving all occurrences of the different OoD objects from the three datasets.

\textbf{Segmentation Evaluation:} 
We follow the standard evaluation protocol for the pixel-level performance measures adopted from \cite{pinggera2016lost,blum2019fishyscapes}.
Namely, these are the Area Under Precision-Recall Curve (AUPRC) and the False Positive Rate at 95\% of True Positive Rate (FPR$_{95}$). 
From a practitioner’s perspective, it is often sufficient only to recognize a fraction of the pixels of an OoD object to detect and localize them.
For evaluating the component-level performance of the OoD segmentation model, the averaged component-wise score $\overline{F}_1$ \cite{chan2021segmentmeifyoucan} serves as our main evaluation metric. We note that the standard evaluation protocol for OoD segmentation only evaluates predictions that fall into the ground truth road regions \cite{pinggera2016lost, chan2021segmentmeifyoucan}. Since expensive ground truth segmentation labels cannot be assumed to be available for large-scale OoD analysis, this assumption must be relaxed for applications that utilize the OoD predictions for downstream tasks. Therefore, we report the $\overline{F}_1$ score on the predicted road regions instead of ground truth road regions.

\textbf{Tracking Evaluation:} We evaluate the object tracker performance using the common multiple object tracking (MOT) metrics \cite{bernardin2008evaluating}. These metrics quantify the algorithm's ability to accurately detect the number of objects present and determine the position of each object. The Multiple Object Tracking Accuracy (MOTA) is a metric that evaluates the tracking algorithm's performance in detecting objects and maintaining their trajectories, regardless of the precision with which the object positions are estimated. On the other hand, the Multiple Object Tracking Precision (MOTP) assesses the tracker's ability to accurately estimate the positions of objects, irrespective of its detection capabilities.

\textbf{Retrieval Evaluation:} To evaluate our retrieval performance, we provide a textual query, in our case, the name of the ground truth classes from the OoD datasets, and we evaluate how well our method succeeds in retrieving the matching OoD road obstacle. We use instance-based precision and recall as metrics. A retrieved instance (such as an image crop supposed to contain an OoD road obstacle object) is called true positive if the majority of the pixels within the corresponding image bounding box semantically belong to the query. Consequently, precision is the fraction of retrieved instances that match the query and recall is the fraction of all instances in the dataset according to the query, which are correctly retrieved. As the retrieval performance depends on a similarity threshold, we report the Precision-Recall Curve for all queries in each dataset.


\subsection{Object-level vs Image-level Processing}
\begin{figure}[!t]
    \centering
    \includegraphics[width=0.99\linewidth]{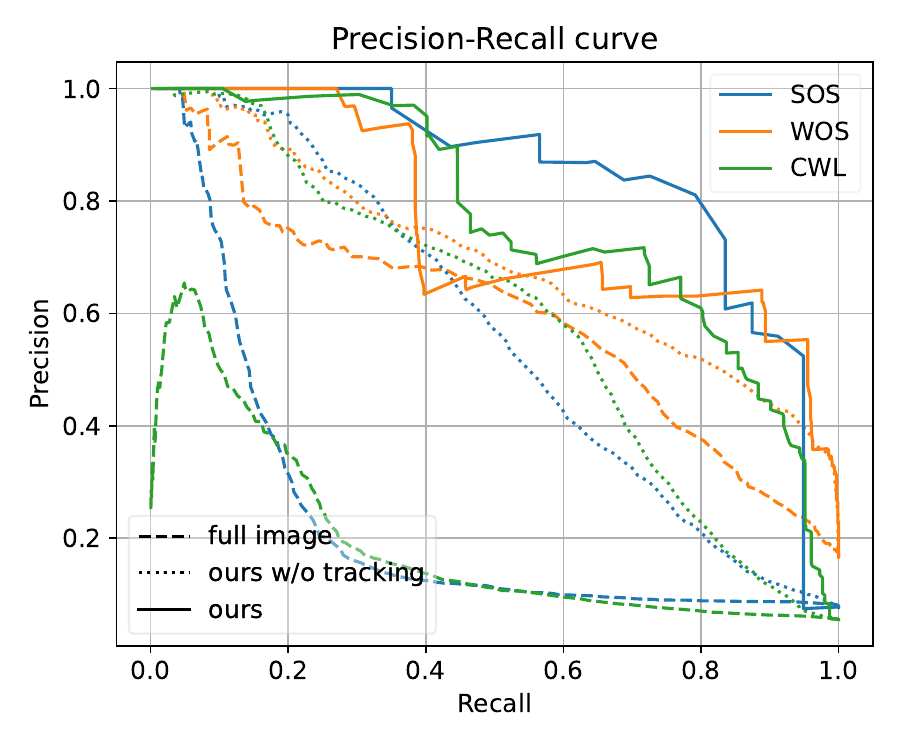}
    \caption{\textbf{Precision-Recall curve for each dataset}: Each curve illustrates the trade-off between precision and recall for varying thresholds. Dashed curves represent the baseline approach of retrieving based on full-scale driving scenes. Solid curves represent our method of retrieving based only on a cut-out of the OoD road obstacle with tracking information, and dotted curves represent our method but without tracking information.
    }
    \label{fig:pr_curve_clip_vs_full_image}
\end{figure}
In the first experiment, we present a comprehensive evaluation of our proposed method for object retrieval. We compare our approach of segmenting, tracking, and embedding cut-outs of OoD road obstacles against the conventional approach of retrieving directly on embeddings of the full image. Our approach is rooted in our observation that object-level information is necessary for retrieving OoD road obstacles in complex driving scenes where OoD road obstacles make up the minority of the full driving scene. Furthermore, we examine the impact of tracking on retrieval performance. For this experiment, we assume optimal conditions where all OoD road obstacles were detected and tracked correctly.

The results for the precision-recall curve for each of the methods and datasets are shown in \Cref{fig:pr_curve_clip_vs_full_image}. The results demonstrate our method's significant advantage in performance over the baseline approach. This is primarily attributed to the fact that OoD road obstacles typically only occupy a minor portion of the overall driving scene. Therefore, relying solely on full-frame retrieval leads to inferior performance. The results also show that tracking plays a role in improving the retrieval results. This is because far-away detections are more challenging to retrieve than closer ones. Therefore, creating a link between detections closer to the camera and far away detections via tracking improves the retrieval performance. 

\subsection{Ablation Study}
\begin{figure}[!t]
    \centering
    \includegraphics[width=0.99\linewidth]{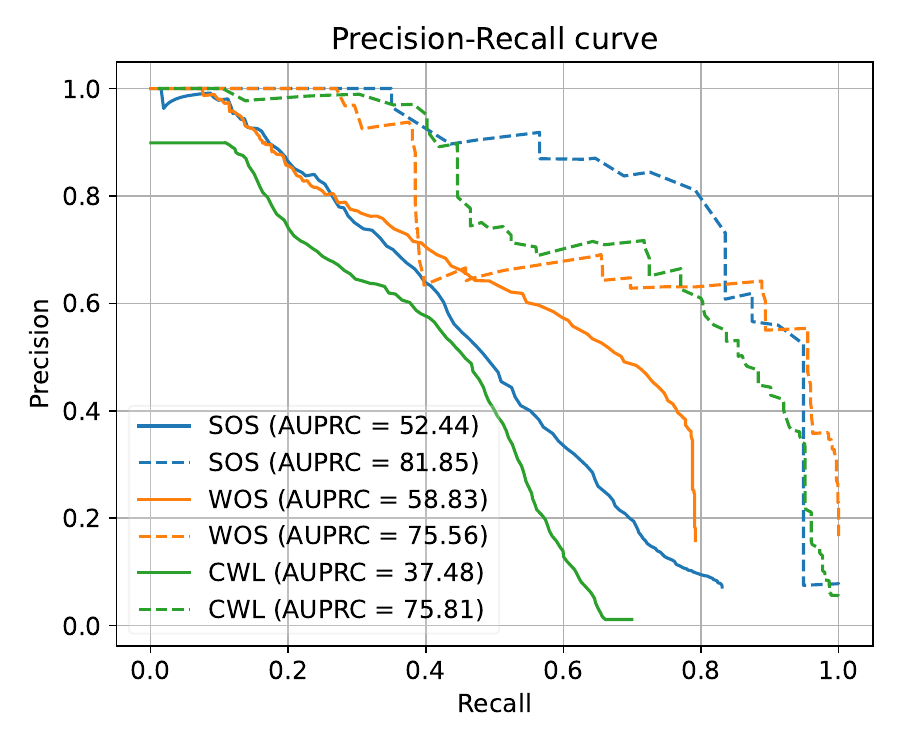}
    \caption{\textbf{Average precision-recall curve for each dataset}: The area under the curve (AUPRC) provides a comprehensive measure of the retrieval performance for each dataset. Solid curves represent our method, where we segment and track the video streams, and dashed curves represent the retrieval performance on ground truth detections and tracking.
    }
    \label{fig:pr_curve_clip_vs_standard}
\end{figure}

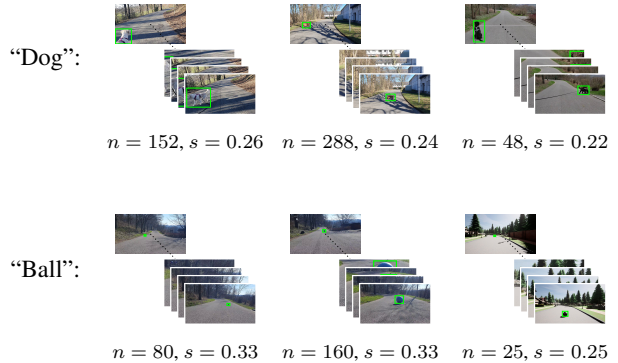
\begin{figure}
    \centering
    \resizebox{\linewidth}{!}{\input{tikz/examples_retrieval}}
    \caption{Examples of retrieved video sequences with the corresponding query, sequence length ($n$), and similarity score ($s$). From left to right, the images correspond to the first and last frame of the sequence.}
    \label{fig:retrieval-examples}
\end{figure}

We conduct an ablation study to understand the contribution and significance of the individual components of our method. In the first experiment, we evaluate the efficacy of our proposed method for the task of OoD road obstacle retrieval. \Cref{fig:pr_curve_clip_vs_standard} shows our retrieval performance measured by the area under the precision-recall curve for the different datasets and compared to the setting with perfect OoD detections.
We note that the tracking performance of the proposed lightweight algorithm is almost perfect when evaluated on ground truth detection.
Therefore, to achieve better trajectories of objects, we require either a more robust tracking method that can compensate for the errors in detections or enhance the segmentation model to reduce false positives. 
We found that, although the performance is reasonable when we assume that all OoD road obstacles are detected, there is still room for improvement. Regarding our OoD segmentation method, despite utilizing a state-of-the-art network, it falls short of capturing all instances of OoD objects (as indicated by dashed curves not reaching a recall value of one).
\Cref{table:full_result} summarizes our evaluation results for OoD segmentation, tracking, and retrieval across three video datasets using different segmentation networks. The table shows a strong correlation between the OoD segmentation network performance ($\overline{F}_1$ score) and the tracking and retrieval performance. This signifies the importance of OoD segmentation for retrieval. \Cref{fig:retrieval-examples} shows qualitative examples of retrieved video sequences.

\begin{table}[!t]
\resizebox{\linewidth}{!}{
\centering
\begin{tabular}{c|c|c|c|c|c|c|c}
    \hline 
    & & \multicolumn{3}{c|}{Segmentation} & \multicolumn{2}{c|}{Tracking} & Retrieval \\
    \hline
    Dataset & Method & AUPRC $\uparrow$ & $ \mathrm{FPR}_{95} \downarrow$ & $\overline{F}_1 \uparrow$ & MOTA $\uparrow$ &  MOTP $\downarrow$  &AUPRC $\uparrow$ \\
    \hline \multirow{2}{*}{SOS} & Entropy max  & 85.20 & 1.30 & 50.40 & 0.32  & 12.45 & 37.53 \\
     & RbA  & \textbf{89.47} & \textbf{0.33} & \textbf{53.58} & \textbf{0.36}  & \textbf{5.93} & \textbf{52.44}\\
    \hline
    \multirow{2}{*}{WOS} & Entropy max  & \textbf{94.92} & \textbf{0.59} & 30.13 & 0.13  & 51.17 & 26.03 \\
     & RbA  & 93.76 & 0.81 & \textbf{48.52} & \textbf{0.23}  & \textbf{16.88} & \textbf{58.83} \\
    \hline
    \multirow{2}{*}{CWL} & Entropy max  & 79.54 & 1.38 & 47.64 & 0.48  & 18.91 & 26.33 \\
     & RbA  & \textbf{86.93} & \textbf{0.59} & \textbf{60.17} & \textbf{0.52}  & \textbf{7.01} & \textbf{37.48}  \\
     \hline
\end{tabular}
}
\caption{OoD object segmentation, tracking, and retrieval results.}
\label{table:full_result}
\end{table}

\begin{table}[!t]
\resizebox{\linewidth}{!}{
\centering
\begin{tabular}{c|c|c|c|c}
    \hline 
    & \multicolumn{1}{c|}{Segmentation} & \multicolumn{2}{c|}{Tracking} & Retrieval \\
    \hline
    Dataset & $\overline{F}_1 \uparrow$ & MOTA $\uparrow$ &  MOTP $\downarrow$  &AUPRC $\uparrow$ \\
    \hline SOS & 68.94 (+15.36) &  0.68 (+0.32)  & 3.17 (-2.76) & 65.01 (+12.57)\\
    WOS & 73.85 (+25.33) &  0.46 (+0.23)  & 7.40 (-9.48) & 64.59 (+5.76) \\
    CWL
     & 63.89 (+3.72) & 0.58 (+0.06)  & 6.62 (-0.39) & 40.87 (+3.39)  \\
    
    \hline
\end{tabular}
}
\caption{OoD object segmentation, tracking, and retrieval results under perfect region of interest, with comparative performance
gains in comparison to RbA in \Cref{table:full_result}.}
\label{table:result_perfect_roi}
\end{table}
\textbf{Detection and Retrieval under Perfect Regions of Interest}:
After analyzing our method, we identify a pattern of multiple false positives occurring on the sidewalk. This can be attributed to the fact that the sidewalk often contains objects that are considered OoD, and our road segmentation and region of interest generation methods are not perfect, sometimes resulting in a region of interest that includes the sidewalk. Therefore, we evaluate our proposed method under perfect regions of interest masks obtained from the ground truth road and OoD road obstacle masks. We evaluate the performance gain in segmentation, tracking, and retrieval.
\Cref{table:result_perfect_roi} shows the results of this experiment and highlights the additive performance gains in comparison to our method. We note that the improvement is due to the decrease in the number of false positive predictions, which leads to better tracking and, therefore, better retrieval scores. We remove the pixel-wise evaluation metrics (AUPRC and $\mathrm{FPR}_{95}$) from the evaluation since these were evaluated using the standard evaluation protocol of \cite{chan2021segmentmeifyoucan}, which is only limited to the ground truth region of interest.  

\begin{table}[!t]
\resizebox{\linewidth}{!}{
\centering
\begin{tabular}{c|c|c|c|c}
    \hline 
    & \multicolumn{1}{c|}{Segmentation} & \multicolumn{2}{c|}{Tracking} & Retrieval \\
    \hline
    Dataset & $\overline{F}_1 \uparrow$ & MOTA $\uparrow$ &  MOTP $\downarrow$  &AUPRC $\uparrow$ \\
    \hline
    SOS & 19.68 (-33.90) &  -1.57 (-1.93)  & 30.41 (+24.48) & 24.29 (-28.15)\\
    WOS & 28.31 (-20.21) &  -0.99 (-1.22)  & 16.66 (+0.22) & 32.55 (-26.28) \\
    CWL & 22.98 (-37.19) & -0.69 (-1.21)  & 14.21 (+7.20) & 35.95 (-1.53)  \\
    \hline
\end{tabular}
}
\caption{OoD object segmentation, tracking, and retrieval results without meta classification, with comparative performance loss in comparison to  RbA method in \Cref{table:full_result}.}
\label{table:result_no_meta}
\end{table}
\textbf{The Effect of Meta Classification}:  
Meta classification \cite{rottmann2019uncertainty, chan2021entropy} poses an additional but negligible computational overhead to the OoD prediction pipeline that significantly reduces false positives. We evaluate the impact of removing Meta Classification from our pipeline. 
\Cref{table:result_no_meta} presents our findings related to the impact of meta-classification on segmentation, tracking, and retrieval performance. As expected, removing meta-classification reduces the  ($\overline{F}_1$ score ) due to increased false positives, leading to worse tracking and retrieval performance.

\section{Conclusion}
This work presents a first approach for the novel task of text-to-video OoD road obstacle retrieval. 
Our primary aim is to address the question of \lq\lq\textit{Have we ever encountered this before?}\rq\rq, a critical question arising during the life cycle of AI components in real-world scenarios of automated driving. 
This would advance the development of automated driving systems by enabling the adaption of driving policies in constantly changing environments.
By leveraging single-frame OoD segmentation, object tracking, and multi-modal embedding of OoD road obstacles, our method provides an effective and efficient solution to retrieve relevant video data in response to AI-related field issues.
The empirical results showcase the clear advantages of our approach of object-level processing over the straight-forward baseline that relies solely on complete image information.
Through exploration of the retrieval task's dependence on segmentation and tracking, we uncover valuable insights into enhancing performance.
Specifically, we note the need for better post-segmentation methods to eliminate false-positive predictions, \ie the prediction of the drivable area as a region of interest for OoD road obstacles and meta classification for automated segment-wise false-positive removal.
We believe this work lays the groundwork for further research into the issue of OoD road obstacle retrieval for a fast response to AI-related safety concerns during deployment.
Thereby, we contribute towards resolving real-world challenges arising in the life cycle of data-driven AI components in automated driving perception systems.

{\small
\bibliographystyle{ieee_fullname}
\bibliography{egbib}
}

\end{document}

%% file: tikz/examples_retrieval.tex
\graphicspath{{figs/experiments}} 

\begin{tikzpicture}

    \node [] () at (-1.5, -.5) {``Dog'':};


    \node [] () at (0, 0) {\includegraphics[width=.12\linewidth]{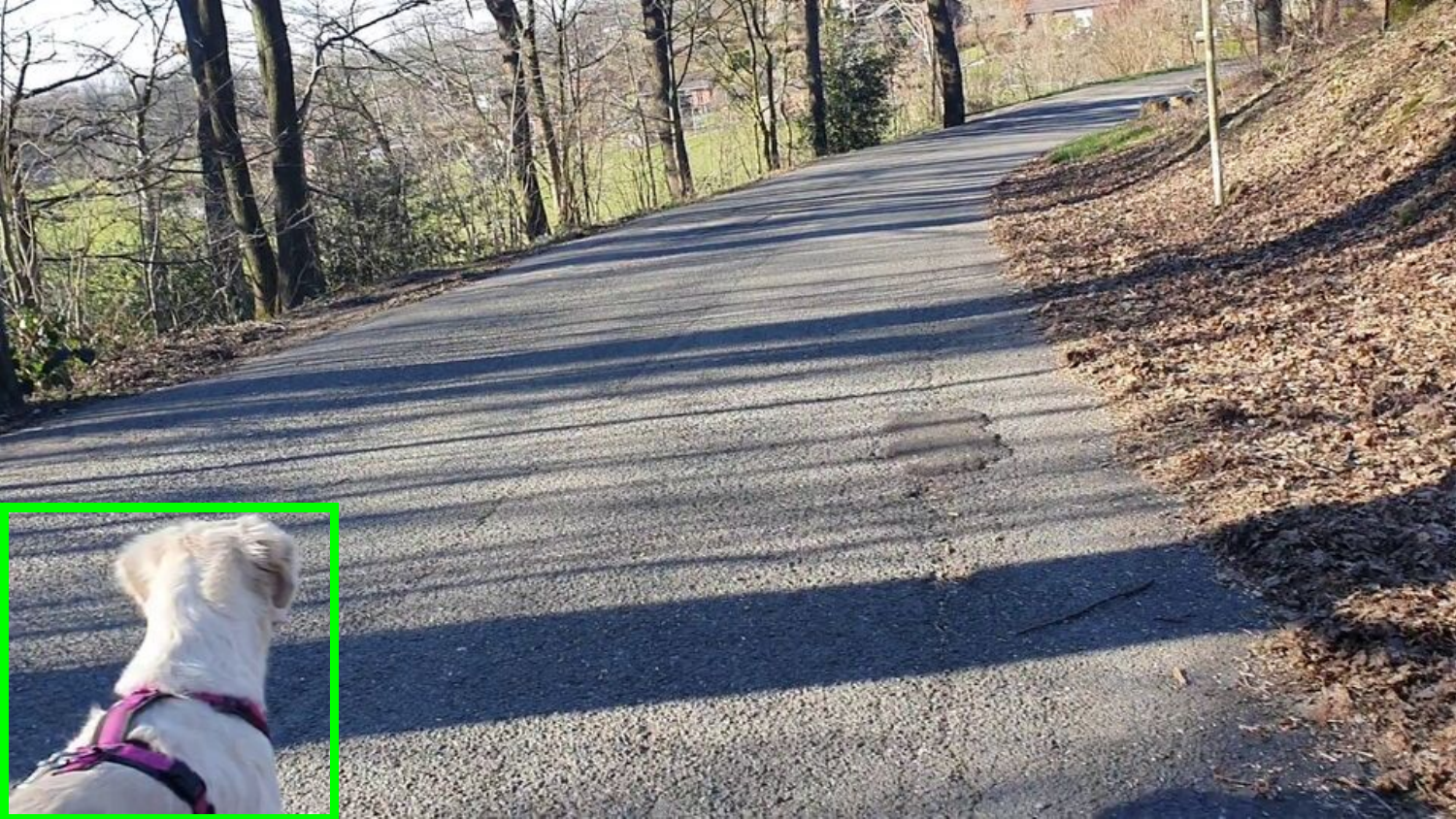}}; 
    \draw [densely dotted] (0,0) -- (1, -1);
    
    \node [] () at (.7, -.7) {\includegraphics[width=.12\linewidth, trim={8cm 4cm 8cm 4cm}, clip, cfbox=white 0.5pt 0pt]{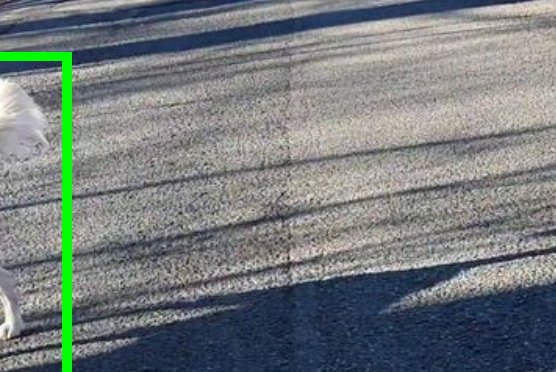}};
    \node [] () at (.8, -.8) {\includegraphics[width=.12\linewidth, trim={6cm 3cm 6cm 3cm}, clip, cfbox=white 0.5pt 0pt]{query_1_seq_1_last.pdf}};
    \node [] () at (.9, -.9) {\includegraphics[width=.12\linewidth, trim={4cm 2cm 4cm 2cm}, clip, cfbox=white 0.5pt 0pt]{query_1_seq_1_last.pdf}};
    \node [] () at (1, -1) {\includegraphics[width=.12\linewidth, cfbox=white 0.5pt 0pt]{query_1_seq_1_last.pdf}}; 

    \node [] () at (.5, -1.7) {\footnotesize $n=152$, $s=0.26$};


    \node [] () at (2.5, 0) {\includegraphics[width=.12\linewidth]{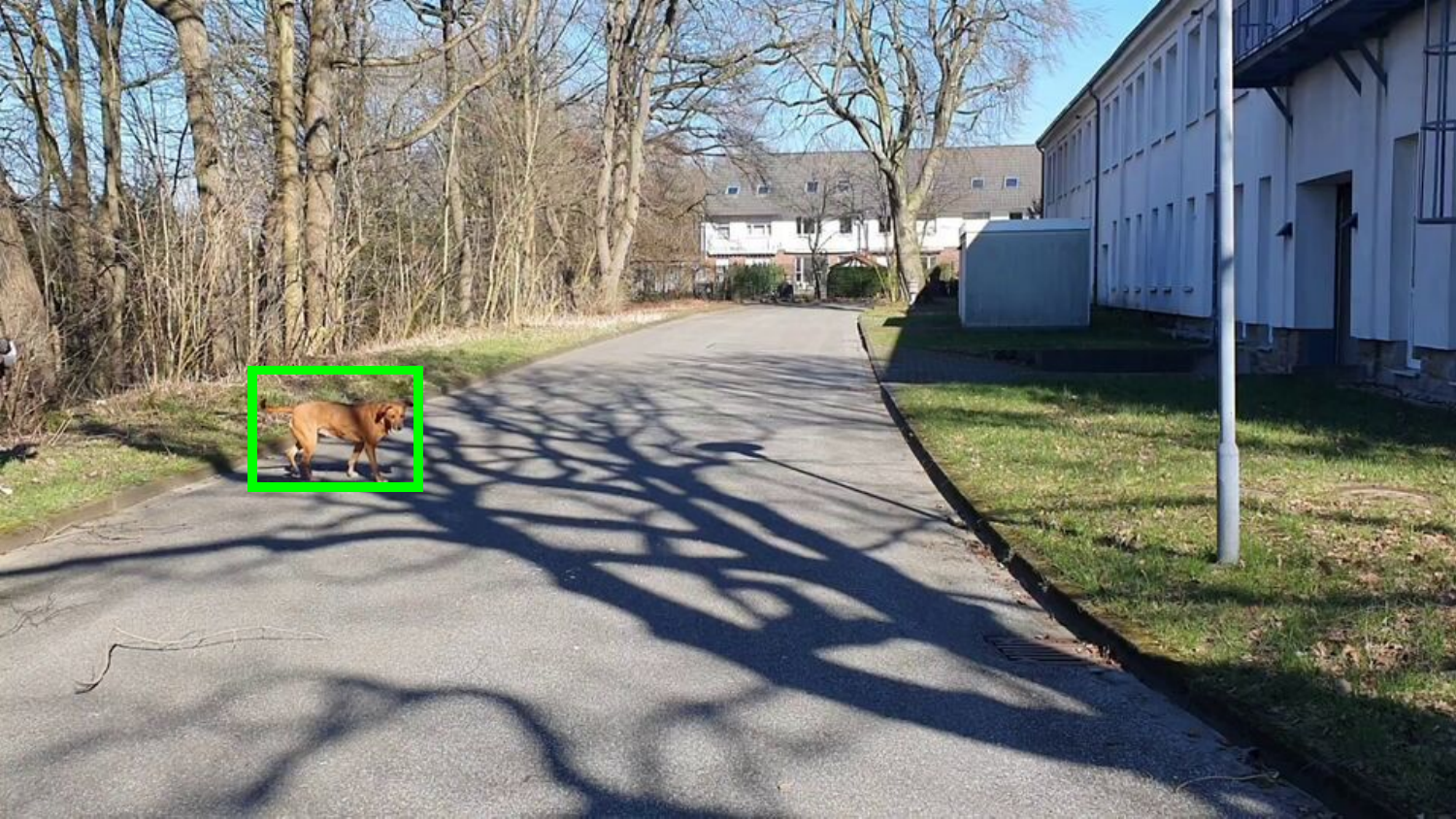}};
    \draw [densely dotted] (2.5,0) -- (3.5, -1);
    \node [] () at (3.2, -.7) {\includegraphics[width=.12\linewidth, trim={8cm 4cm 8cm 4cm}, clip, cfbox=white 0.5pt 0pt]{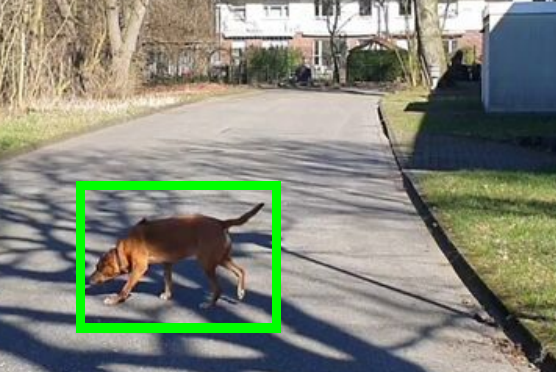}};
    \node [] () at (3.3, -.8) {\includegraphics[width=.12\linewidth, trim={6cm 3cm 6cm 3cm}, clip, cfbox=white 0.5pt 0pt]{query_1_seq_2_last.pdf}};
    \node [] () at (3.4, -.9) {\includegraphics[width=.12\linewidth, trim={4cm 2cm 4cm 2cm}, clip, cfbox=white 0.5pt 0pt]{query_1_seq_2_last.pdf}};

    \node [] () at (3.5, -1) {\includegraphics[width=.12\linewidth, cfbox=white 0.5pt 0pt]{query_1_seq_2_last.pdf}};

    \node [] () at (3., -1.7) {\footnotesize $n=288$, $s=0.24$};


    \node [] () at (5, 0) {\includegraphics[width=.12\linewidth]{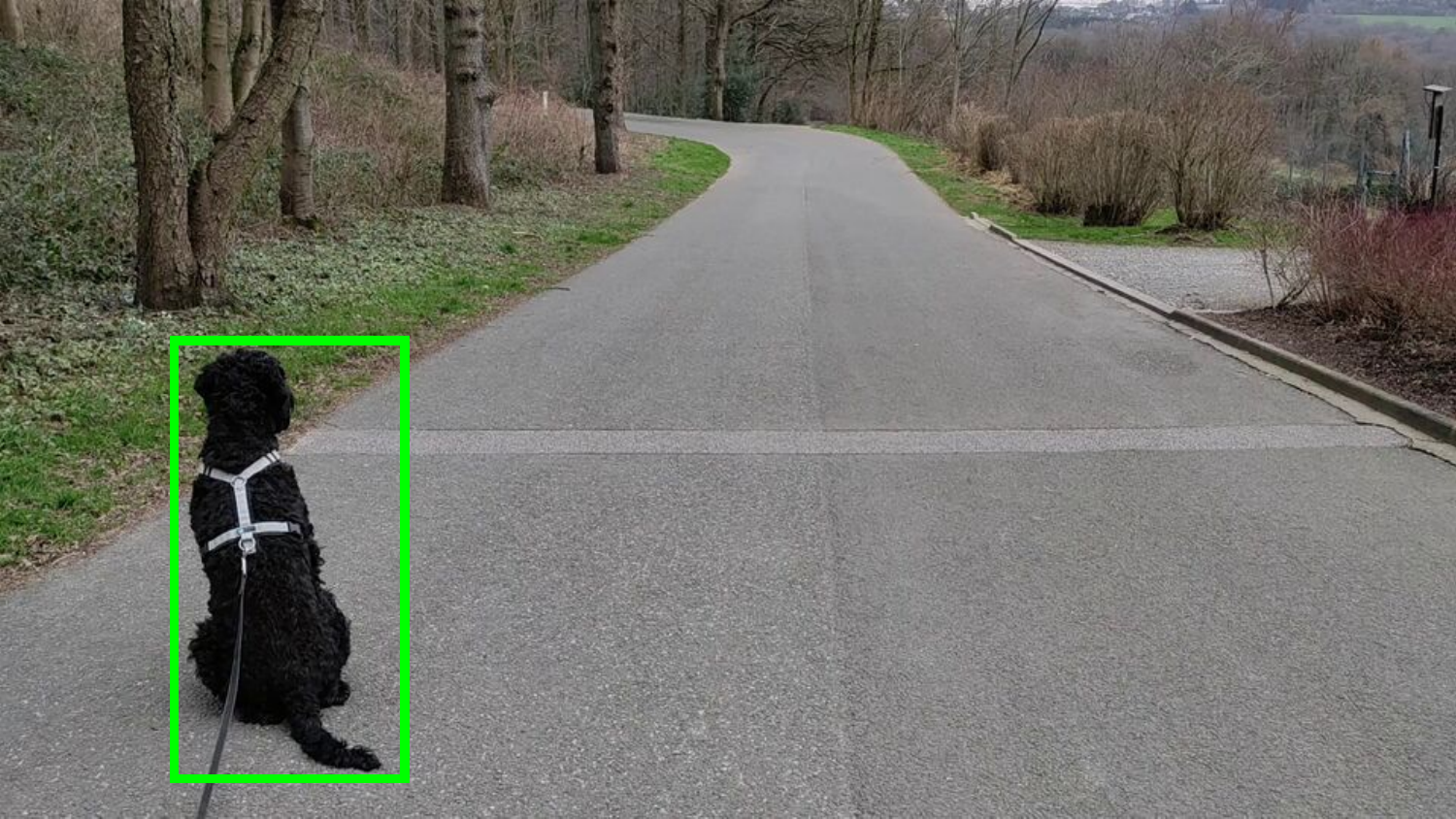}};
    \draw [densely dotted] (5,0) -- (6, -1);
    \node [] () at (5.7, -.7) {\includegraphics[width=.12\linewidth, trim={8cm 4cm 8cm 4cm}, clip, cfbox=white 0.5pt 0pt]{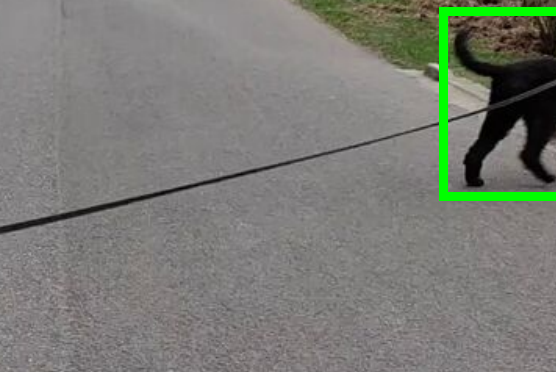}};
    \node [] () at (5.8, -.8) {\includegraphics[width=.12\linewidth, trim={6cm 3cm 6cm 3cm}, clip, cfbox=white 0.5pt 0pt]{query_1_seq_3_last.pdf}};
    \node [] () at (5.9, -.9) {\includegraphics[width=.12\linewidth, trim={4cm 2cm 4cm 2cm}, clip, cfbox=white 0.5pt 0pt]{query_1_seq_3_last.pdf}};

    \node [] () at (6, -1) {\includegraphics[width=.12\linewidth, cfbox=white 0.5pt 0pt]{query_1_seq_3_last.pdf}};

    \node [] () at (5.5, -1.7) {\footnotesize $n=48$, $s=0.22$};

    \node [] () at (-1.5, -3.5) {``Ball'':};


    \node [] () at (0, -3) {\includegraphics[width=.12\linewidth]{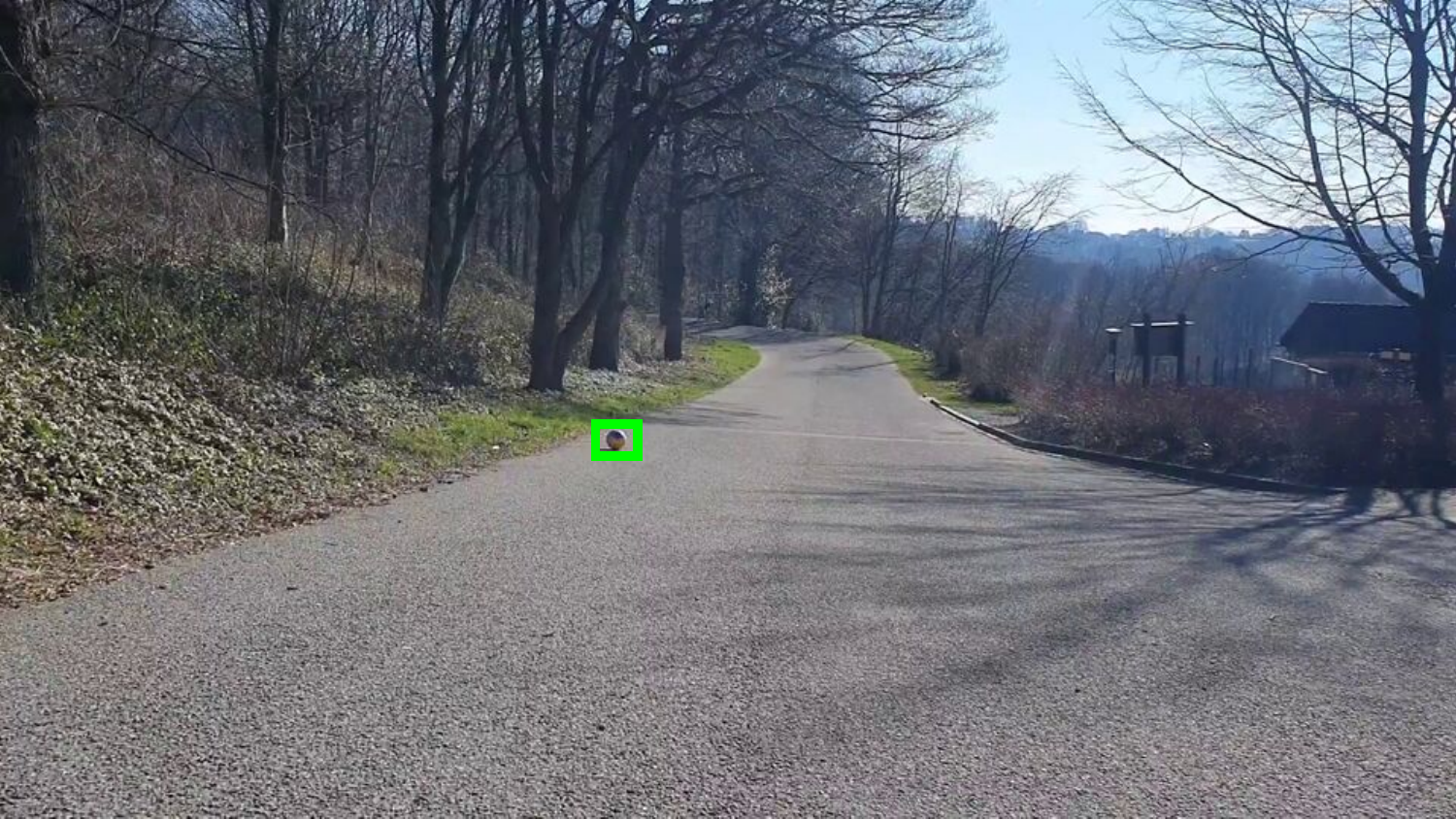}};
    \draw [densely dotted] (0,-3) -- (1, -4);
    \node [] () at (.7, -3.7) {\includegraphics[width=.12\linewidth, trim={8cm 4cm 8cm 4cm}, clip, cfbox=white 0.5pt 0pt]{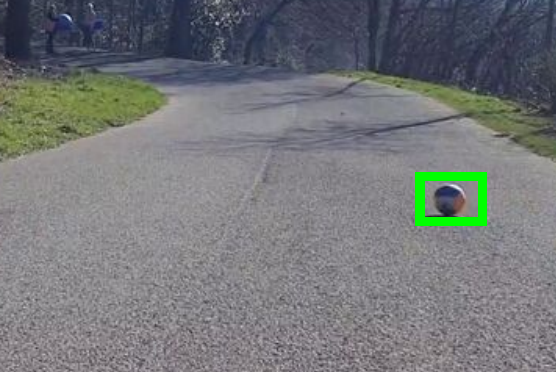}};
    \node [] () at (.8, -3.8) {\includegraphics[width=.12\linewidth, trim={6cm 3cm 6cm 3cm}, clip, cfbox=white 0.5pt 0pt]{query_2_seq_1_last.pdf}};
    \node [] () at (.9, -3.9) {\includegraphics[width=.12\linewidth, trim={4cm 2cm 4cm 2cm}, clip, cfbox=white 0.5pt 0pt]{query_2_seq_1_last.pdf}};
    
    \node [] () at (1, -4) {\includegraphics[width=.12\linewidth, cfbox=white 0.5pt 0pt]{query_2_seq_1_last.pdf}};

    \node [] () at (.5, -4.7) {\footnotesize $n=80$, $s=0.33$};


    \node [] () at (2.5, -3) {\includegraphics[width=.12\linewidth]{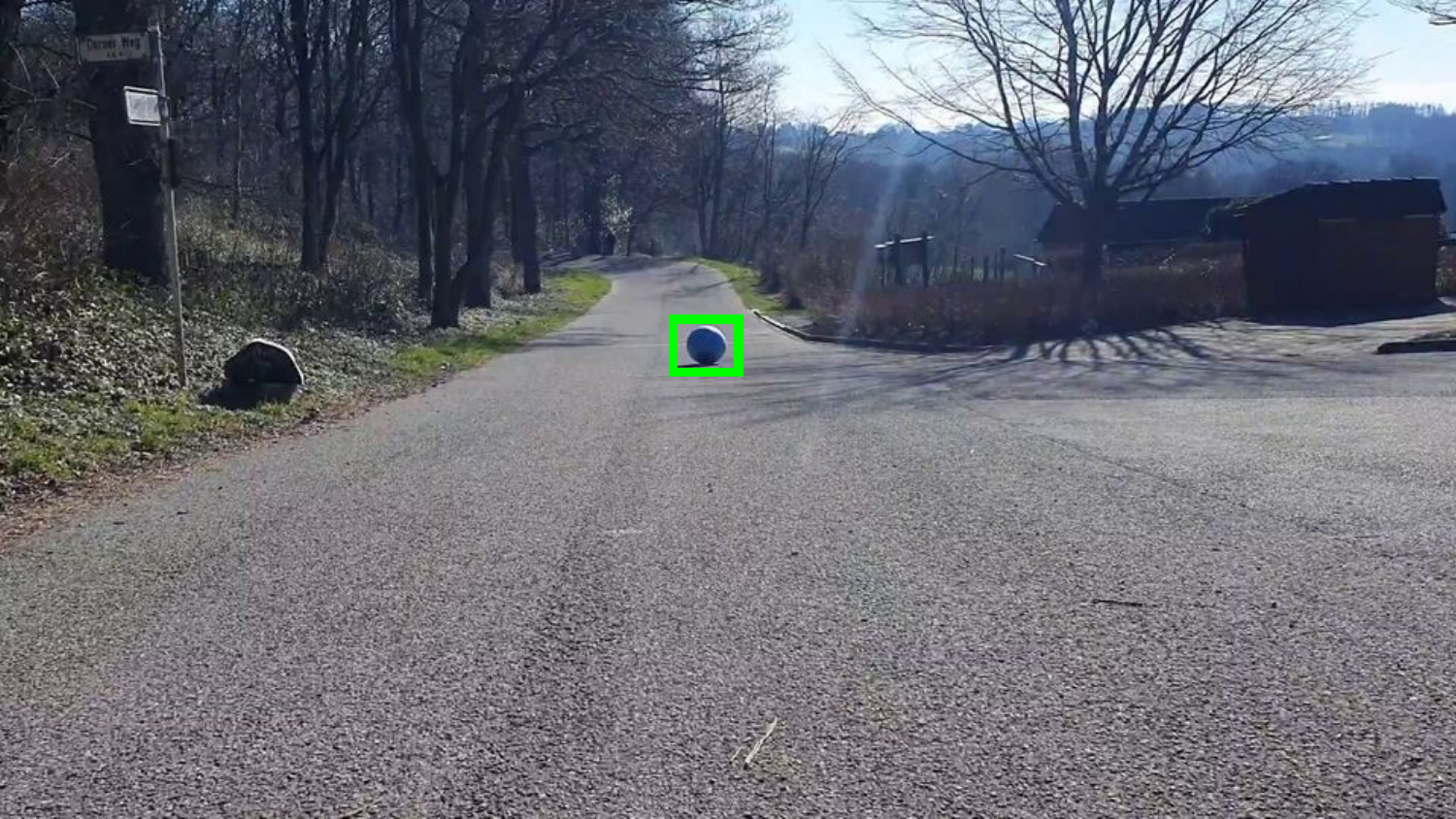}};
    \draw [densely dotted] (2.5,-3) -- (3.5, -4);
    \node [] () at (3.2, -3.7) {\includegraphics[width=.12\linewidth, trim={8cm 4cm 8cm 4cm}, clip, cfbox=white 0.5pt 0pt]{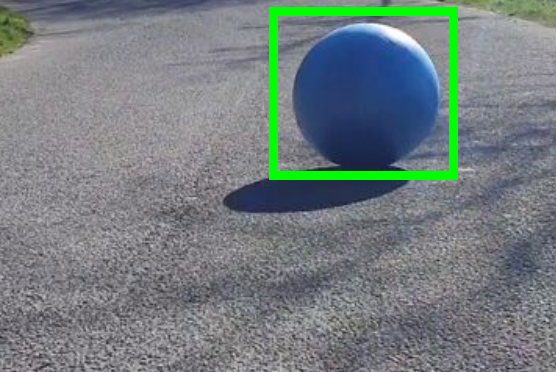}};
    \node [] () at (3.3, -3.8) {\includegraphics[width=.12\linewidth, trim={6cm 3cm 6cm 3cm}, clip, cfbox=white 0.5pt 0pt]{query_2_seq_2_last.pdf}};
    \node [] () at (3.4, -3.9) {\includegraphics[width=.12\linewidth, trim={4cm 2cm 4cm 2cm}, clip, cfbox=white 0.5pt 0pt]{query_2_seq_2_last.pdf}};
    
    \node [] () at (3.5, -4) {\includegraphics[width=.12\linewidth, cfbox=white 0.5pt 0pt]{query_2_seq_2_last.pdf}};

    \node [] () at (3., -4.7) {\footnotesize $n=160$, $s=0.33$};


    \node [] () at (5, -3) {\includegraphics[width=.12\linewidth]{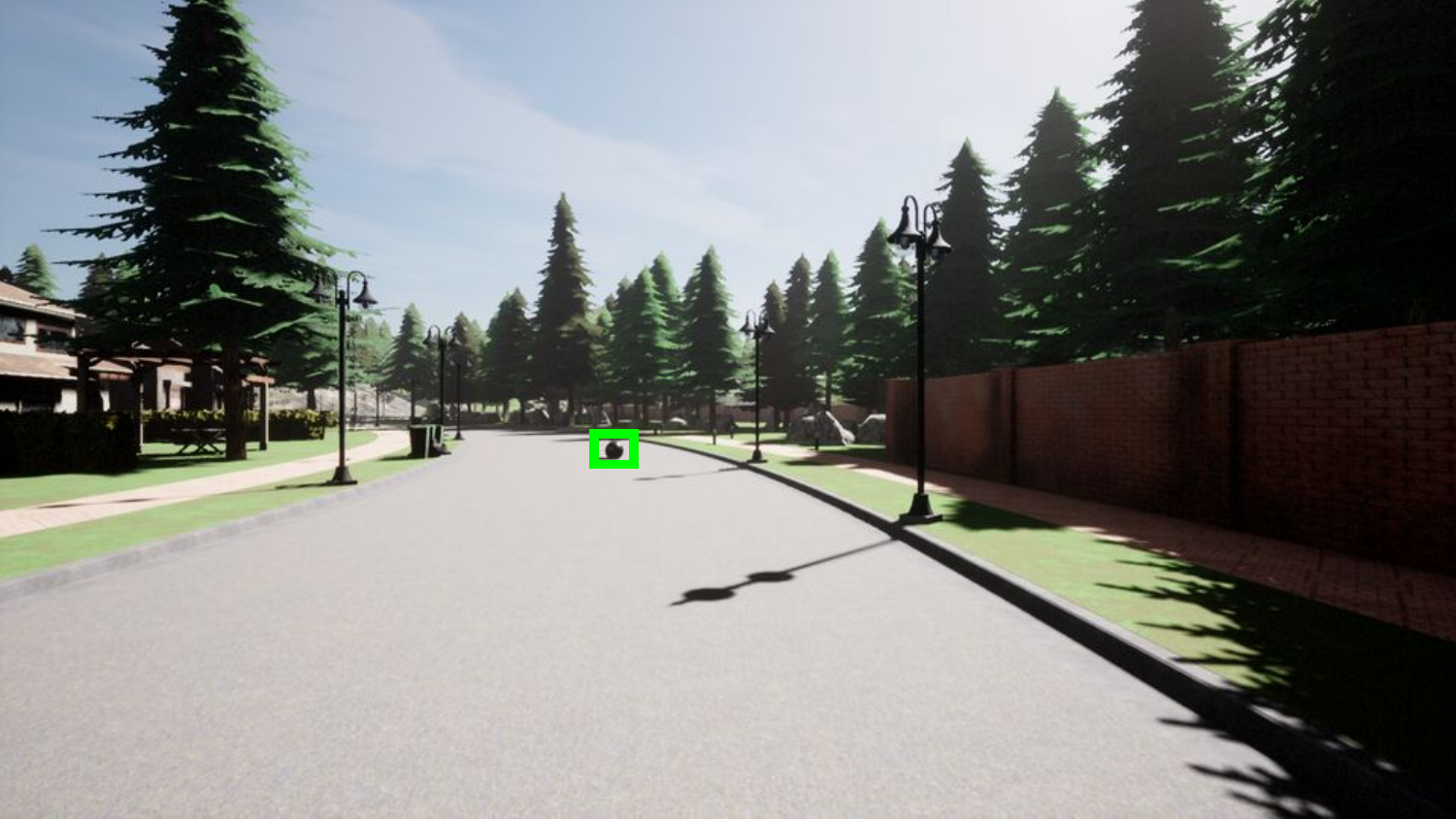}};
    \draw [densely dotted] (5,-3) -- (6, -4);
    \node [] () at (5.7, -3.7) {\includegraphics[width=.12\linewidth, trim={8cm 4cm 8cm 4cm}, clip, cfbox=white 0.5pt 0pt]{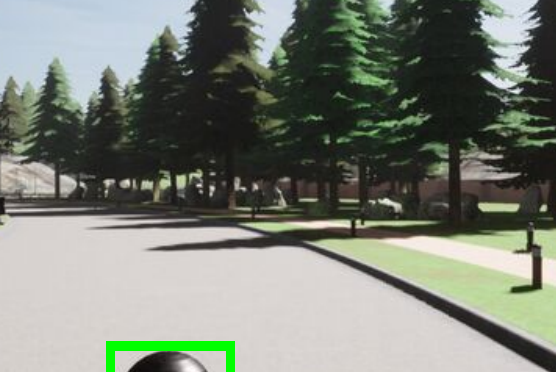}};
    \node [] () at (5.8, -3.8) {\includegraphics[width=.12\linewidth, trim={6cm 3cm 6cm 3cm}, clip, cfbox=white 0.5pt 0pt]{query_2_seq_3_last.pdf}};
    \node [] () at (5.9, -3.9) {\includegraphics[width=.12\linewidth, trim={4cm 2cm 4cm 2cm}, clip, cfbox=white 0.5pt 0pt]{query_2_seq_3_last.pdf}};
    
    \node [] () at (6, -4) {\includegraphics[width=.12\linewidth, cfbox=white 0.5pt 0pt]{query_2_seq_3_last.pdf}};
    
    \node [] () at (5.5, -4.7) {\footnotesize $n=25$, $s=0.25$};

\end{tikzpicture}